\documentclass[letterpaper, 11pt]{IEEEtran} 
\usepackage{subcaption} 

\usepackage{graphicx} 
\usepackage[square, numbers]{natbib}  
\usepackage{algorithm}
\usepackage{algorithmic}
\usepackage[utf8]{inputenc} 
\usepackage[T1]{fontenc}    
\usepackage{booktabs}       
\usepackage{amsfonts}       
\usepackage{amsmath}       
\usepackage{amssymb} 

\usepackage{nicefrac}       
\usepackage{microtype}      
\usepackage{xcolor}         
\usepackage{multirow}
\usepackage{array}
\usepackage{newfloat}
\floatstyle{ruled}
\renewcommand{\thetable}{\arabic{table}}

\usepackage{titlesec}
\titleformat*{\section}{\Large\bfseries}
\titleformat*{\subsection}{\large\bfseries}

\begin{document}
\pagestyle{headings}

\title{DMS: Differentiable Mean Shift for Dataset Agnostic Task Specific Clustering Using Side Information} 

\author{
    Michael Hobley, Victor Prisacariu \\ Active Vision Laboratory\\
  University of Oxford\\
  \texttt{[mahobley, victor]@robots.ox.ac.uk} \\
}

\maketitle

\begin{abstract}
We present a novel approach, in which we learn to cluster data directly from side information, in the form of a small set of pairwise examples. Unlike previous methods, with or without side information, we do not need to know the number of clusters, their centers or any kind of distance metric for similarity. Our method is able to divide the same data points in various ways dependant on the needs of a specific task, defined by the side information. Contrastingly, other work generally finds only the intrinsic, most obvious, clusters. Inspired by the mean shift algorithm, we implement our new clustering approach using a custom iterative neural network to create Differentiable Mean Shift (DMS), a state of the art, dataset agnostic, clustering method. We found that it was possible to train a strong cluster definition without enforcing a constraint that each cluster must be presented during training. DMS outperforms current methods in both the intrinsic and non-intrinsic dataset tasks.
\end{abstract}

\begin{figure*} [t]
 \centering
            \includegraphics[width=0.9\linewidth]{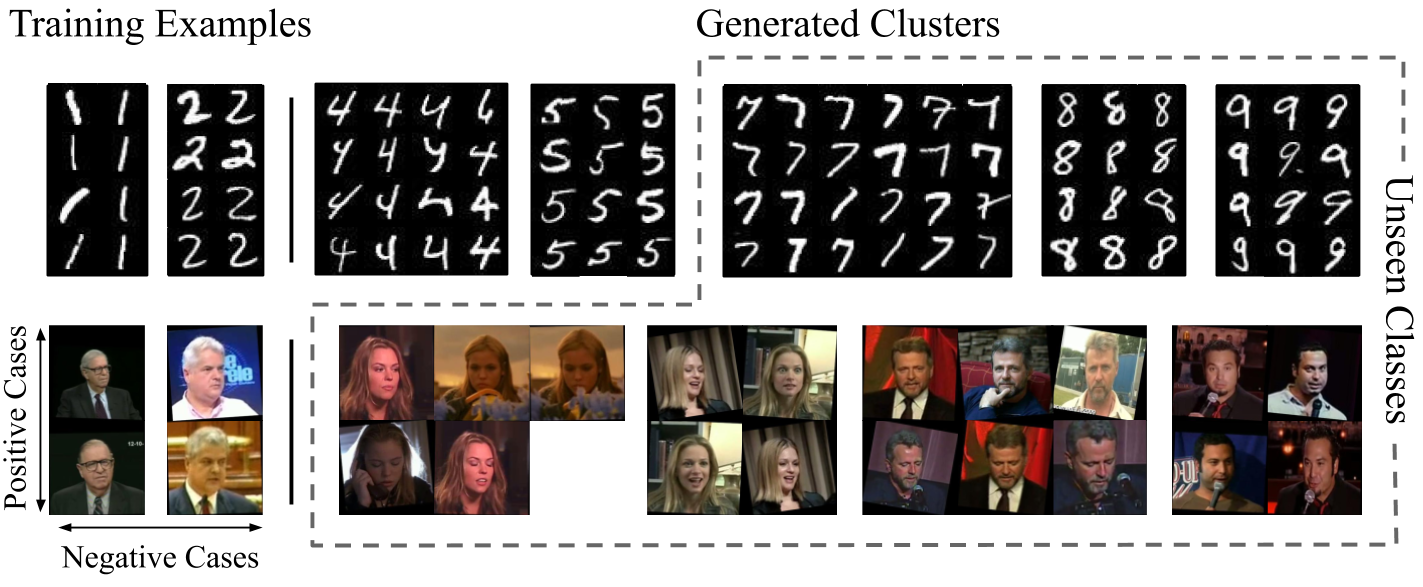}
             \caption{Examples of learnt clusters given side information.
             Positive examples contain two data points known to be from the same class and negative examples are from different classes, here the same class is either the same digit (top) or same person (bottom) and negative examples are of two different classes. Our method learns to correctly identify unseen classes, for example the 7, 8, 9 clusters are not present when training on MNIST (top).}
             \vspace{-5mm}
    \label{fig:teaser}
 \end{figure*}  
\section{Introduction}
Clustering algorithms have been extensively studied as they are a ubiquitous tool.  
Classical clustering approaches normally require no known ground truth class labels but do need either a known number of clusters as in $k$-means \cite{Lloyd1982} or DBSCAN \cite{Sander1998}, or  a predefined similarity distance tolerance, as in mean shift \cite{Fukunaga1975}.  
In many cases it is difficult to specify an informative distance or to know the number of clusters in the dataset.
Generally deep learning based methods use these classical algorithms either to supervise their training \cite{Shah2018,shah2018deep} or to directly cluster a projected feature space \cite{Chang2017,Yang2017}. These approaches suffer from the same problems as the classical methods, and require parameter tuning.
Another, often ignored difficulty is the fact that many datasets could be clustered in multiple valid ways, as seen in Figure \ref{fig:intrisic_non_intrinsic}, each useful for a different end task.

Our approach is inspired by other contemporary work \cite{Xing,han2020automatically,Han} that uses pairwise similarity based side information to cluster a dataset, and by mean shift which iteratively finds clusters using a distance metric as a proxy for similarity; thus, we constructed Differentiable Mean Shift (DMS).
DMS is the natural progression of these two approaches as it removes the proxy and trains a similarity kernel to find clusters directly from only side information. As in other contemporary work \cite{Xing,han2020automatically,Han}, the side information takes the form of pairwise 'similar/dissimilar' labels. With DMS we iteratively train a kernel to evaluate the similarity of a given pair of points in the context of a specific task, thus DMS can cluster the same dataset in multiple valid ways by altering the side information.
Unlike other classical or deep clustering methods \cite{Shah2018,shah2018deep,Chang2017,Dizaji2017}, this learnt definition 
shows promise without the need for: i) wider knowledge of the dataset (for example, the number of clusters, their centers or a distance metric), ii) all classes to be present or evenly represented in the example data iii) an embedded feature space or an explicit dimensional reduction stage, iv) hyper-parameter tuning for different datasets or tasks, which are present in other methods \cite{Shah2018,shah2018deep,Chang2017,Dizaji2017,Hsu2015,Liu2015,Ceccarelli2008,Li2018}. 
As we can learn a task specific similarity kernel without all classes present in the side information DMS is able to identify novel, unseen classes at inference time as seen in Figure \ref{fig:teaser}.
As our method operates as a single forward pass through a lightweight network which does not require tuning, dimensional reduction or an understanding of the whole dataset, it is trivial to add it into another deep learning method either as a head or for end-to-end training.

Our contributions are (i) a new approach to clustering, which directly clusters points trained using only side information rather than using distance tolerances or a number of clusters (ii) a differentiable, mean shift inspired framework that can be easily added to other deep networks (iii) a state of the art algorithm that outperforms other contemporary methods on intrinsic tasks and is capable of performing non-intrinsic task clustering on various datasets.

The remainder of this paper is structured as follows. Section \ref{relatedwork} briefly discusses classical clustering algorithms before discussing current deep learning based approaches to clustering data.
Section \ref{meanshift} outlines and formalises the mathematics of the classical mean shift algorithm, and our differentiable mean shift process. Section \ref{experiments} discusses the details of our method and evaluates it in the context of other recent clustering  approaches.

\section{Related Work}\label{relatedwork}
Here we consider classical approaches to clustering before discussing recent deep learning based approaches.

\subsection{Classical Clustering}
Classic deterministic clustering algorithms can be roughly grouped into spatial clustering \cite{Lloyd1982,Sander1998} and spectral clustering \cite{VonLuxburg2007}. 

Spatial clustering methods use a metric function, typically the L1 norm, Euclidean norm or cosine distance, to evaluate the difference or similarity between two data points.
For instance $k$-means \cite{Lloyd1982}, minimises the sum of the distances between each data point and its closest cluster centers. $K$-means++ \cite{Arthur2007} improves the initialisation by finding spread out sample points, $k$ points are randomly selected, but the selection is weighted based on the squared distance between it and other already sampled points. Another improvement on $k$-means came from Xing et al. \cite{Xing}, who focused on learning a more appropriate distance metric using pairwise side information. This idea has since been extensively explored \cite{Ceccarelli2008,ailon2018approximate,choudhury2019top,kim2017relaxed,mazumdar2017clustering}.
$K$-means and its variants benefit from being both simple and robust, however, in practice, $k$ is often unknown. 

Another branch of spatial methods adopts a bottom up strategy, in which small clusters are merged using a metric function until only one cluster is left. Examples are the Ward's hierarchical clustering (AC-W) \cite{Murtagh2014} and DBSCAN \cite{Sander1998}. DBSCAN repeatedly expands a cluster from a seed if the density is above a threshold area. The GDL \cite{Zhang2012} agglomeratively merges sub clusters according to a metric function that takes into account local geometry using in/out degrees of a data point. 

Spectral clustering methods \cite{VonLuxburg2007,Nie2011,Shi} convert a data point graph into the eigenspace of its laplacian.
Given $k$ is equal to the multiplicity of the eigenvalues and clusters can be retrieved from the eigenvectors, the eigenspace can be used to improve the $k$-means initialisation. 

Classic clustering approaches use simple metric functions and work well with low-dimension data. However, they generalise poorly in higher dimension space. This is known as the curse of dimensionality \cite{Aggarwal2001,Domingos:2012:FUT:2347736.2347755,Bellman1962}. 

\subsection{Deep Clustering}\label{rel_deep}
Current deep learning methods attempt to either (i) project features into a space that can be clustered using classical methods effectively (ii) mimic conventional spatial/spectral clustering approaches with a set of trainable layers
\cite{DBLP:journals/corr/abs-1801-07648}.

For example, Deep Clustering Network (DCN) \cite{Yang2017} uses a regularised $k$-means loss. JULE \cite{Yang2016} uses an agglomerative clustering loss to learn feature embeddings.
Deep Embedding for Clustering (DEC) \cite{Xie2016} and DEPICT \cite{Dizaji2017} both use an assignment loss which measures the KL divergence between their predictions and a soft assignment.
Robust Continuous Clustering (RCC) \cite{Shah2018} and Deep Continuous Clustering (DCC) \cite{shah2018deep} use a locality-preserving loss to learn a robustified feature embedding that is consistent with a classical $k$-NN graph. 
DAC \cite{Chang2017} iteratively improves the feature space driving similar points together using binary pairwise-classification loss.

In order to tackle the problem posed by high dimensional spaces, many methods have applied deep learning to project the features into a lower dimensional space. 
For instance, 
RCC-DR \cite{Shah2018} and DCN both train a linear autoencoder for dimension reduction, 
DEPICT and DCC both improve upon previous works by adding deep autoencoders.
While autoencoders 
are the most commonly used dimensional reduction approach, other methods have been proposed to similar effect, for example Hu et al. \cite{Hu2017}, Hsu and Lin \cite{Hsu2018}, and \cite{VittalPremachandran2017}, reduce their feature spaces using self-augmentation and a GAN respectively.

The impact of an improved feature space is clear, DAC \cite{Chang2017} generates a feature space that can be effectively clustered by $k$-means even for challenging datasets, however, DAC offers negligible improvements to the clustering itself. In this work we are focusing on clustering a feature space without encoding it, as it is not always possible or efficient to train an encoder for the clustering, for example when the clustering is added as a head. Of course the above encoding methods could compliment our approach well.

Deep clustering approaches that use classical clustering methods as a base have the same flaws as those classical methods i.e. they require some supervision in the form of dataset-specific parameters. This parameter tuning significantly decreases the usefulness of these clustering approaches for end-to-end training, as well as for non deep-learning applications.

Recently there have been various works that focus on semi or completely supervised clustering \cite{Hsu2015,Fogel2019,Kong2018,unknown}. These approaches however have generally focused on learning a feature encoding optimised for clustering rather than focusing on creating a robust clustering method to work on any features. As discussed above, these works could be used to compliment our work but focus on a fundamentally different aspect of the problem. 

Following Xing et al. \cite{Xing}, the use of pairwise side information for deep clustering has been further explored as an alternative to direct supervision \cite{han2020automatically,Liu2015,Hsu2015,Li2018,Mazumdar2017}. However, these approaches have similar drawbacks to previous clustering methods. They require a knowledge of the data beyond solely pairwise examples, commonly the number of ground truth clusters present in the dataset \cite{han2020automatically,Hsu2015,Liu2015}.

\section{Mean Shift}\label{meanshift}

Given a set of data points, mean shift \cite{Fukunaga1975} finds the appropriate cluster centers and compares these in order to cluster the data points. Cluster centers are found by iteratively averaging the position of points currently believed to be inliers to the same cluster.
In classical mean shift this is based around two assumptions: firstly, points that are similar belong to the same cluster, and, secondly, that euclidean distance is a good proxy for similarity.
The mean shift algorithm is formulated as follows.

Given a set, $\mathcal X$,
where $|\mathcal{X}|=M$ and any data point $\mathbf x\in\mathcal{X}$ is a $N$-Dimension vector $\mathbb{R}^N$.
The sample mean function, $m:\mathcal{X} \times \mathbb R^{N} \rightarrow \mathbb R^N$, on $\mathcal X$  is defined by:
    \begin{equation}\label{eq:mean_shift}
        \bar {\mathbf x}^{(n+1)} = m(\mathcal{X}, \bar {\mathbf x}^{(n)}) = \frac{\sum_{\mathbf x \in \mathcal X} \mathbf x K(\mathbf x, \bar {\mathbf x}^{(n)}) }{\sum_{\mathbf x \in \mathcal X} K(\mathbf x, \bar{\mathbf x}^{(n)})},
    \end{equation}
where $\bar{\mathbf x}^{(n)}$ is the current sample mean calculated in the $n$-th iteration, $\mathbf x \in \mathcal X$ is a data point, and $K(.)$ is the kernel function which evaluates the similarity of two points, $K:\mathbb R^N \times \mathbb R^{N}\rightarrow \{0,1\}$. 

Given the sample mean function $m(.)$ defined on some collection of data with an initial value $\bar{\mathbf{x}}^{(0)}$, a cluster centre together with the cluster inlier data points can be calculated using the mean shift algorithm~\cite{Fukunaga1975}. The sample mean is iteratively updated using Equation \ref{eq:mean_shift} until the distance of sample means between two consecutive iterations becomes sufficiently small, i.e. $\|\bar{\mathbf x}^{(n+1)} - \bar{\mathbf x}^{(n)} \| < \tau $, where $\tau$ is the threshold. The difference between the converged sample mean $\bar{\mathbf x}$ and the start point $\bar{\mathbf x}^{(0)}$ is known as the {\it mean shift}.

\subsection{Differentiable Kernel}\label{MS_diff_kernel}

The original Mean shift algorithm relies on a kernel function $K_F(.)$ measuring the similarity between two data points using their Euclidean distance, which is known as the flat kernel, and is not differentiable. Any data point within the $\lambda\text{-ball} \subset \mathbb R^N$ centred at $\bar{\mathbf{x}}$ of the cluster is considered as an inlier.
    \begin{equation}
        K_F(\mathbf x_1, \mathbf x_2)=
        \begin{cases}
          1 & \text{if}\ \|\mathbf x_1 - \mathbf x_2\|_2 \leqslant \lambda \\
          0 & \text{if}\ \|\mathbf x_1 - \mathbf x_2\|_2 \geqslant \lambda
        \end{cases}
    \end{equation}

Another commonly used similarity kernel is the unit Gaussian Kernel
 \cite{Cheng1995}. The Gaussian Kernel
has no neighbourhood characteristic, including all points in $\mathcal X$, but scales the importance of points based on their distance to the sample mean. While the Gaussian Kernel is differentiable, its  representation capability is limited.

Instead of using distance as a proxy for similarity we propose a differentiable kernel based on a neural network, $K_D: \mathbb R^N\times\mathbb R^{N} \rightarrow [0,1]$, that learns similarity directly. 
The differentiable kernel, $K_D(.)$ can produce a probability to measure the similarity between two data points or a data point and the sample mean $\bar {\mathbf x}$. Particularly, the more similar, the output of $K_D(.)$ is closer to 1. Due to the flexibility of the neural network, $K_D(.)$ can be much more descriptive than the classical analytical metrics, such as the aforementioned flat or Gaussian kernels. 

In high dimensional space it is desirable to be able to act non-symmetrically towards a dataset's dimensions based on a task's requirements. 
A neural network is ideal for this as each dimension can be acted on independently to find those most salient to a given task. 
In contrast simpler metrics like the Euclidean norm are not able to apply the same contextual knowledge of the feature space, and always evaluate dimensions with greater absolute variation as more significant.
This neural network ability decreases the need for explicit dimensional reduction and removes the problems of poor representation inherent in such a reduction.
The context based dimensional independence also allows a single feature space to be clustered differently dependent on task.
The specific architectures of our differentiable kernels can be seen in Table \ref{table:Architecture}.
The output of $K_D$ is bounded to $[0,1]$ by a sigmoid function. This sigmoid also introduces a non-linear scaling similar to the traditional Gaussian kernel, this scaling increases and decreases the impacts of the areas close to and far from the sample mean respectively.
  \begin{table}
          \centering
          \caption{Differential subtraction and concatenation based kernel architectures, as described in Section \ref{ablation}.}
          \renewcommand{\arraystretch}{1.2}
\begin{tabular}{|c|c|c|}
\hline
\multirow{2}{*}{\begin{tabular}[c]{@{}c@{}} Inputs \end{tabular}}  
& Sample Mean, $\bar{\mathbf{x}}_i$  & N \\ \cline{2-3}
  & Data point, $\mathbf{x}$ &     N \\ \hline 
\end{tabular}

\vspace{0.25em}

\begin{tabular}{|c|c|c|}
 \cline{1-3} 
\multirow{4}{*}{\begin{tabular}[c]{@{}c@{}} \\ \\ $K_D(\mathbf{x}, \bar{\mathbf{x}}_i)$ \\ $subtract$ \end{tabular}} & Subtract, $(\mathbf{x} - \bar{\mathbf{x}}_i)$ &    N  \\ 
 \cline{2-3}
 
 & 
 \begin{tabular}[c]{@{}c@{}}
 $
     \begin{bmatrix} 
        \text{fully connected,}  & N/2 \\
        \text{fully connected,}  & N/4 \\
        \text{fully connected,}  & 1 
    \end{bmatrix} $ \\
    Sigmoid 
    \end{tabular} &      
    
    \begin{tabular}[c]{@{}c@{}}
$   
        \begin{matrix} 
         N/2 \\
        N/4 \\
         1  \\
         1 \\
    \end{matrix} $ \\
      \end{tabular}
\\
 \cline{1-3}
\multirow{4}{*}{\begin{tabular}[c]{@{}c@{}} \\ \\ $K_{D}(\mathbf{x}, \bar{\mathbf{x}}_i)$ \\ $concat$ \end{tabular}}
 & Concatenate, $concat(\mathbf{x}, \bar{\mathbf{x}}_i)$ &      2 $\times$ N \\ 
 \cline{2-3} 
 
 & 
 \begin{tabular}[c]{@{}c@{}}$
        \begin{bmatrix} 
        \text{convolution,} &2  \times  1, N \\
        \text{fully connected,}  &N/2 \\
        \text{fully connected,} &N/4 \\
        \text{fully connected,}  &1 
    \end{bmatrix} $ \\
    Sigmoid
    \end{tabular} &  \begin{tabular}[c]{@{}c@{}}
$
    \begin{matrix} 
         N/2 \\
        N/4 \\
         1  \\
         1 \\
         1 \\
    \end{matrix} $ \\
  
    \end{tabular}\\ 
     \cline{1-3} 
    
\end{tabular}

            \label{table:Architecture_sub}
        \label{table:Architecture}
    \end{table}

\subsection{Differentiable Mean Shift}
\begin{figure*}[t] 
    \centering
                   \begin{subfigure}{0.32\linewidth}
            \centering
            \includegraphics[width=\linewidth]{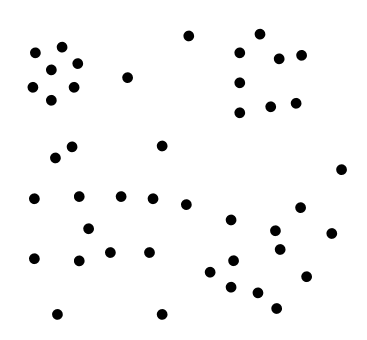} 
            \caption*{Data Points}
            \end{subfigure}
            \begin{subfigure}{0.32\linewidth}
            \centering
            \includegraphics[width=\linewidth]{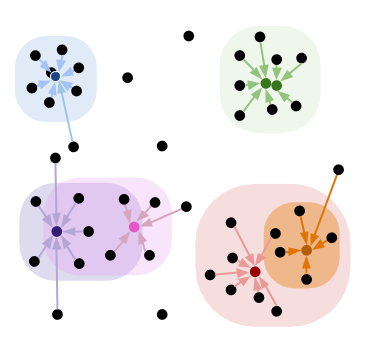} 
            \caption*{Center Finding Stage }
            \end{subfigure}
                \begin{subfigure}{0.32\linewidth}
            \centering
            \includegraphics[width=\linewidth]{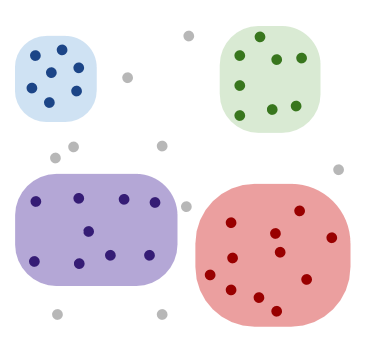} 
            \caption*{Inlier Prediction Stage}
            \end{subfigure}
          \caption{Algorithm Stages. First, the centers  are located with an initial approximation of the clusters. Then, given the centers inliers are then found, and the clusters are refined, similar clusters are combined, subset clusters are incorporated and noise points are removed.}
        \label{fig:process}
    \end{figure*}

\begin{figure}
        \centering
        \includegraphics[width=\linewidth]{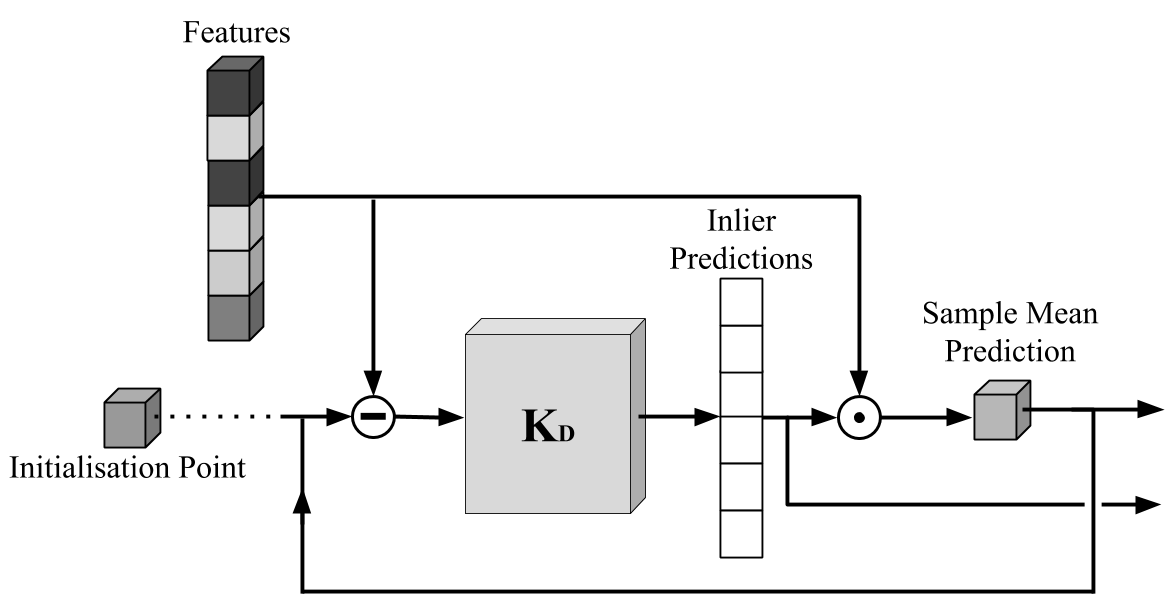}
        \caption{The DMS framework for cluster prediction.}
    \label{fig:loop_diagram}
 \end{figure}

Our method, which in practicality is a single pass through a neural network, is comprised of two stages, iterative center finding and refined inlier prediction, as shown in Figure \ref{fig:process}.
During the first stage the cluster centers are found by iteratively taking the weighted average of their inliers, using their predictions, $K_D(\mathbf x_i, \bar {\mathbf x})$, as the weighting. In the second stage the confidence of each point being an inlier to each cluster is used to combine similar centers, this stage also finds the true cluster each point is associated with.
The second stage can also be thought of as the same as a final iteration of the first stage; it uses the same network and weights to find inlier predictions, but does not subsequently use these predictions as weights to find a center.
To estimate the sample mean vector of one cluster, we initialise $\bar{\mathbf x}^{(0)} = \mathbf x_i$, where $\mathbf{x}_i \in \mathcal X$, we then predict the sample mean iteratively, $\bar{\mathbf{x}}^{(n+1)} = m(\mathcal X, \bar{\mathbf{x}}^{(n)})$, where $n$ is the iteration step index, one step can be seen in Figure \ref{fig:loop_diagram}.

Unlike classical mean shift, we do not determine convergence by defining a distance tolerance between the centers found in consecutive iterations. Instead, we use the prediction of inliers given the current and previous sample mean. If two consecutive sample means, $\bar{\mathbf{x}}^{(n-1)}$ and  $\bar{\mathbf{x}}^{(n)}$ predict the same points to be inliers and outliers, the algorithm has converged and the cluster center has been found.
In this way, we can calculate $M$ sample means, $\bar{\mathbf{x}}_i$, initialised from $\mathbf{x}_i$, where $i=1,..,M$.

We evaluate sample mean similarity in the same way as we find our iterative convergence. This is due to the fact that while we observe that various samples, $\mathbf x_i$, may robustly converge to an identical cluster center, i.e. $\|\bar{\mathbf{x}}_i-\bar{\mathbf{x}}_j\|_2 < \epsilon$ where $\epsilon$ is an infinitesimally small value; we also observe that many samples that belong to the same ground truth cluster may result in different but similar sample means. 
We define two centers to be the same if they predict the same points to be inliers and outliers with similar confidences.

Formally, we calculate a confidence matrix $\mathbf H\in\mathbb [0,1]^{M\times M}$ where the element $\mathbf H_{r,c}$ at the $r^{th}$ row and the $c^{th}$ column represents the confidence that the $c^{th}$ data point belongs to the $r^{th}$ candidate cluster $\mathbf H_{r,c}=K_D(\mathbf{x}_c, \bar{\mathbf{x}}_r)$.

A similarity matrix over the candidate sample means can be calculated $\mathbf S=\tilde{\mathbf H} \tilde{\mathbf H}^\top$, where $\tilde{\mathbf H}$ is the bi-stochastic normalisation of $\mathbf H$. Each element $\mathbf S_{i,j}$ measures the similarity between the $i^{th}$ and $j^{th}$ sample means. The final unique cluster centers can be found by binarising $\mathbf S$ and extracting connected components.
The same binarising tolerance of $0.5$ is used for all datasets. However, we found that any value between about $0.1$ and $0.95$ will provide similar results as the values are driven to either $1$ or $0$ by the training loss, see Equation \ref{eq_BCE}.
Each data point is then assigned to the cluster with the highest inlier confidence for that point.

It may occur that a point that originally converged to a particular center is not assigned to the associated cluster during the final stage either because a different center is more confident of its inliership or because no cluster is confident of its inliership. 
This is generally a positive effect as points not associated with any cluster or on the border of two clusters might converge to an incorrect center during our first stage, due to the lack of a hard neighbourhood constraint in $K_D$.
The clusters could then be further refined in various ways. For example, trivial solutions could be removed if you know this case is not found in the data. we do not make this assumption during our testing.

It is worth noting that unlike classical mean shift our algorithm does not need to be initialised from every point in the dataset. Classical mean shift classifies a point as belonging to a cluster if it converges to the associated center;
our algorithm only needs to be initialised from one point per cluster, this point locates the cluster's center, our second stage then finds all inliers of this cluster.

\subsection{Training}
During training we generate data as follows.
A given dataset contains a set of categories, $\mathcal C$, and each category $c\in\mathcal C$ contains a number of samples, $|c|$, this cardinality  may vary between categories.
One training instance is composed of: a randomly picked data point called the initialisation point, $(c^+,i)$, where $c^+\in\mathcal C$ and $i$ is the index of the initialisation point within the category, a set of positive samples, $\mathcal P$, a set of negative samples, $\mathcal N$, and a set of unlabelled samples,  $\mathcal U$. We use the sets $\mathcal{P}$ and $\mathcal{N}$ to define our side information labels of `similar' and `dissimilar' respectively.

The category of the initialisation point, $c^+$, is named the positive category. The positive samples are randomly drawn from the points who's side information indicates a similarity with the initialisation point, i.e. 
$\mathcal P=\{(c^+,j)\}$. The negative samples are selected from the rest of the points with side information, i.e. $\mathcal N=\{(c^{-},
i)|c^-\neq c^+\cap c^-\in \mathcal C\}$,
and the unlabelled samples $\mathcal U$ are drawn from all points with and without side information, $\mathcal U=\{(c, i)|c \in \mathcal C\}$. 

The specific class label, $c^+$, is only used to creating the side information in the form of pairwise relational labels. 
The use of side information instead of the specific class label during training allows the clustering network to learn a class independent representation of similarity.

The initialisation point, $\mathcal P$ and $\mathcal N$ are all taken from a training set with side information, a small randomly drawn subset of the complete dataset.  All points in the training set were excluded during testing.
The unlabelled samples, $\mathcal U$, are taken from the entire dataset, and are excluded from the training loss.
The unlabelled data points are only added to increase the cardinality and variance of the clusters during training.
All classes and data points in $\mathcal P$, $\mathcal N$ and $\mathcal U$ are randomly and independently selected. Points can be repeated in a training instance, 
this allows for training with data where there are not enough available data points
to create the correct positive class size.

During training we use four mean shift iterations, the impact of this is further discussed in Section \ref{ablation}.
We used the Binary Cross Entropy 
of the inlier predictions and the side information of the labelled points as the training loss, see Equation \ref{eq_BCE}. Predictions of the unlabelled points are not used in the loss. 
    \begin{equation}\label{eq_BCE}
    loss = -\sum_{i \in{\mathcal P,\, \mathcal N}} (y_i\log(p_i)+(1-y_i)\log(1-p_i)) 
    \end{equation}
where $y_i = 1$ if $x_i \in \mathcal P$ and $y_i = 0$ if $x_i \in \mathcal N$, and $p_i$ is the prediction confidence that $x_i$ is in $\mathcal{P}$. This loss consistently pushes predictions to the extremes of the unit interval,
with values between $0.05$ and $0.95$ rarely predicted after training.
\section{Experiments}\label{experiments}
In this section we evaluate DMS using standard datasets and metrics for both intrinsic and non-intrinsic tasks, before discussing our architectural and training decision. 

\subsection{Datasets and Metrics}
We draw datasets from various domains, including: object detection, facial recognition, handwriting comprehension and text classification based on word frequency. To create a fair comparison, we used the standard benchmark descriptors and reduced datasets where appropriate, as in the experimental section of \cite{shah2018deep}.
Specifically, we show results of experiments conducted on MNIST \cite{LeCun1998},  COIL100 \cite{Nene1996}, YouTube Faces (YTF) \cite{Wolf2011} and RCV1 \cite{Lewis2004}.

\medskip
We report results using the standard metrics: clustering accuracy (ACC) \cite{Yang2010} and Normalised Mutual Information (NMI) \cite{Strehl2002}.  Vinh et al. \cite{Vinh:2010:ITM:1756006.1953024} reported that NMI is biased towards the generation of finer grained clusters, and that ACC is unreliable in datasets where there is significant variation in cluster cardinality. For this reason we also include Adjusted Mutual Information (AMI) \cite{Vinh:2010:ITM:1756006.1953024} in all results tables.
ACC, NMI and AMI are defined by Equation \ref{ACC}, 5 and 6
respectively, and are all in the range $[0,1]$, where $1$ indicates ideal performance.
    \begin{equation}\label{ACC}
        \text{ACC}(\mathbf{l}, \hat{\mathbf{l}}) = \max_{map} \frac{\sum_{i \in{M}} \delta(\hat{l}_i,  map(l_i) )}{M}
    \end{equation}
    where $\mathbf{l}$ and $\hat{\mathbf{l}}$ are the predictions and ground truth labels, respectively, $\delta(x, y) = 1$ if $x = y$,  ($\delta(x, y) = 0$, otherwise), and $map$ ranges over all one-to-one mappings between predicted and ground truth labels, efficiently found through use of the Hungarian algorithm \cite{Kuhn1955}.
    \begin{equation}\label{NMI}
        \text{NMI}(\mathbf{l}, \hat{\mathbf{l}}) = \frac{I(\mathbf{l}, \hat{\mathbf{l}})}{\sqrt{H(\mathbf{l})H(\hat{\mathbf{l}})}}
    \end{equation}
    \begin{equation}\label{AMI}
        \text{AMI}(\mathbf{l}, \hat{\mathbf{l}}) = \frac{I(\mathbf{l}, \hat{\mathbf{l}}) - E[I(\mathbf{l, \hat{l}})]}{\sqrt{H(\mathbf{l})H(\hat{\mathbf{l}})} - E[I(\mathbf{l},\mathbf{\hat{l}})]}
    \end{equation}

where $ I(\cdot,\cdot)$ is the mutual information between its inputs, $H(\cdot)$ is the entropy and $E[\cdot]$ is the expectation function. 

\subsection{Experiment Training}
During training, 5\% of the total dataset had pairwise side information.
The total number of data points used in a training instance, $|\mathcal P|+|\mathcal N|+|\mathcal U|$, is a random integer between 200 and 300 and varies between different batches. 
The ratio of points with to without side information in a training instance, 
$|\mathcal P \cup \mathcal N| : |\mathcal U|$, varies between $[\frac{1}{3}, 1]$ 
according to the uniform distribution; 
the ratio of positively to negative pairs, $|\mathcal P|:|\mathcal P \cup \mathcal N|$, varies between $[\frac{1}{20}, \frac{1}{10}]$ according to the uniform distribution. 
For datasets with 10 or fewer classes with similar cardinalities this means the positive set is under-represented during training, and for classes with more than 20 classes, the positive set is over-represented. This did not appear to have any significant effect on performance. Each batch is comprised of 96 training instances.

We found that for datasets where clusters were similar,
i.e. cardinality, inter-cluster variance and intra-cluster variance were comparable, a significantly smaller amount of training data could be used with no detriment to the performance.
COIL100 for instance could be trained with side information present for only 0.7\% of the data, approximately one point per class, from each of 25 of the total 100 clusters, without significant performance differences to using the entire training set. This shows the networks ability to generalise and identify unseen classes.
Conversely, when there are few and unique clusters with regard to cardinality and variance it is important that each cluster is well represented in the training data.

\subsection{Results}\label{testandresults}

In the ideal case we would initialise our method from all data points in the test set at inference time. However we found that as long it was likely a point was initialised in each cluster there was no detriment to performance from reducing the number of initialisation points. Due to computational constraints we initialise from 500 data points randomly selected from the test set. We did not ensure that each ground truth cluster contained an initialisation point explicitly.
We conducted mean shift iterations which included all points in the test dataset from these initialisation points, until the centers converged. 

We recognise that in many cases
there are numerous useful ways to divide a dataset.
Other methods generally only find the most obvious clusters, the `intrinsic' task, while some do allow alteration, this takes the form of abstract parameter tuning in terms of the number or variance of clusters. In contrast our method is versatile and unbiased towards any task and significantly out performs other methods on less obvious, `non-intrinsic', tasks, as seen in Table \ref{table:redefined_task}.

\begin{figure*}
 \centering
            \includegraphics[width=\linewidth]{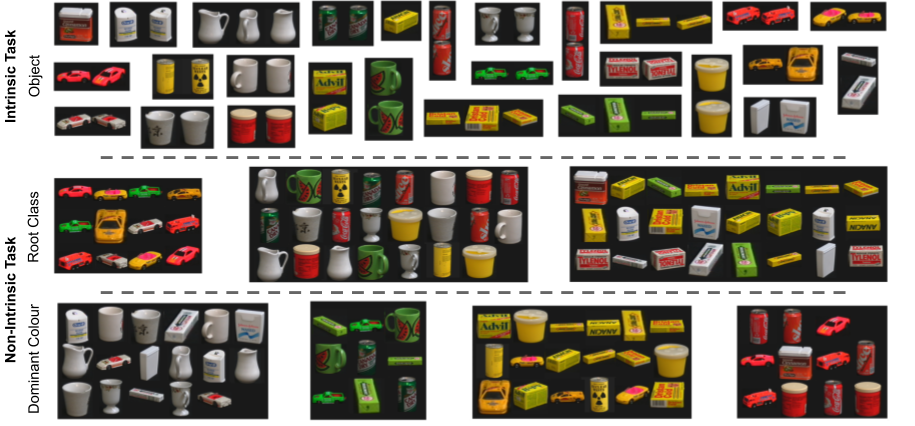}
             \caption{Different clustering tasks using the same data points. The intrinsic task is the most `obvious' divisions of a dataset this is the task other approaches generally solve. 
             As our method is example lead it has no bias towards the intrinsic task over other problems.}
    \label{fig:intrisic_non_intrinsic}
 \end{figure*}

\noindent\textbf{Intrinsic tasks} are those which are most obvious given all facets of the data. In all of our datasets the intrinsic task involves clustering based on a clear object class, reflecting the way a person would likely cluster the data points first, seen in the top row of Figure \ref{fig:intrisic_non_intrinsic}. 
We report our methods performance on the intrinsic tasks of each dataset in Table \ref{table:results}. 
We understand that this comparison is somewhat unfair due to our use of a small amount of side information, however we include these numbers for context.

\noindent\textbf{Non-intrinsic tasks} are the less obvious task, that generally require some alterations to importance of some facets of a dataset. These tasks can be either hierarchical or orthogonal to the intrinsic task or somewhere in between. A hierarchical task is one where multiple complete classes from the intrinsic task are grouped together to form `parent classes'. For example in COIL100 we test on root object classes and dominant object colour, examples are shown in the second and third rows of Figure \ref{fig:intrisic_non_intrinsic}. The root object classes group all object classes
into larger `parent' object-type classes, for example all toy cars are combined. An orthogonal task is one in which every intrinsic cluster contains one instance from the new task. For example, the images of cars from COIL100 could be separated based on orientation (agnostic to the specific object), this is orthogonal to separating the specific objects (agnostic to the orientation).

As other approaches generally find the divisions associated with the intrinsic task of a dataset, orthogonal tasks are performed worse than chance, and it would be unfair to compare these results.
We report the performance of DMS and other contemporary methods on two hierarchical non-intrinsic tasks in Table \ref{table:redefined_task}, we adjusted the hyper-parameters of the benchmark methods to maximise their chance of success on each of the non-intrinsic tasks.
  \begin{table*}
    \centering
        \caption{The ability of our method and other contemporary methods to learn various clustering tasks from the same COIL100 features. 
        Both tasks are using the same features as the intrinsic task, shown in Table \ref{table:results},
        but focus on a different aspect or scale of the semantic information.}
        
\setlength\tabcolsep{4.5pt}
\renewcommand{\arraystretch}{1.2}

 

\begin{tabular}{|c|ccc|ccc|ccc|}
\hline
 & \multicolumn{3}{c|}{\textbf{Intrinsic Task} } & \multicolumn{3}{c|}{\textbf{Root Class} } & \multicolumn{3}{c|}{\textbf{Dominant Colour}}\\
 
Algorithms & \textbf{ACC} & \textbf{NMI} & \textbf{AMI} & \textbf{ACC} & \textbf{NMI} & \textbf{AMI} & \textbf{ACC} & \textbf{NMI} & \textbf{AMI} \\
\hline
RCC-DR & 0.471 & 0.544 & 0.461 & 0.825 & 0.963 & 0.957 & 0.415 & 0.476 & 0.359 \\
DCC & 0.858 & 0.967 & 0.962 & 0.417 & 0.486 & 0.365 & 0.484 & 0.547 & 0.480 \\ \hline 
DMS (sub) & \textbf{0.990} & \textbf{0.996} & \textbf{0.994} & \textbf{0.990} & \textbf{0.997} & \textbf{0.996} & \textbf{0.817} & \textbf{0.790} & \textbf{0.751} \\ \hline 
\end{tabular}
    \label{table:redefined_task}
    \end{table*}
    \begin{table*}
        \centering
        \caption{The clustering accuracy of our approach compared against 14 baselines on the intrinsic task of each dataset. Accuracy is measured using the standard accuracy metric (ACC), Normalised Mutual Information (NMI), and Adjusted Mutual Information (AMI).
        Results other than our own are as found in \cite{shah2018deep}.}
\begin{tabular}{|c|ccc|ccc|ccc|ccc|}	

\hline	
\addlinespace[0.1em]
& \multicolumn{3}{c|}{\textbf{MNIST}  } & \multicolumn{3}{c|}{\textbf{COIL100}  } & \multicolumn{3}{c|}{\textbf{YTF} } & \multicolumn{3}{c|}{\textbf{RCV1}} \\

Algorithms	&	\textbf{ACC}	&	\textbf{NMI}	&	\textbf{AMI}	&	\textbf{ACC}	&	\textbf{NMI}	&	\textbf{AMI}	&	\textbf{ACC}	&	\textbf{NMI}	&	\textbf{AMI}	&	\textbf{ACC}	&	\textbf{NMI}	&	\textbf{AMI}	\\	\hline	
\addlinespace[0.1em]
K-means++	&	0.532	&	0.5	&	0.5	&	0.621	&	0.835	&	0.803	&	0.624	&	0.788	&	0.783	&	0.529	&	0.355	&	0.355	\\																					
AC-W	&	0.571	&	0.679	&	0.679	&	0.697	&	0.876	&	0.853	&	0.647	&	0.806	&	0.801	&	0.554	&	0.364	&	0.364	\\																					
DBSCAN	&	0.000	&	0.000	&	0.000	&	0.921	&	0.458	&	0.399	&	0.675	&	0.756	&	0.739	&	0.571	&	0.017	&	0.014	\\																					
SEC	&	0.545	&	0.469	&	0.469	&	0.648	&	0.872	&	0.849	&	0.562	&	0.760	&	0.745	&	0.425	&	0.069	&	0.069	\\																					
LDMGI	&	0.723	&	0.761	&	0.761	&	0.763	&	0.906	&	0.888	&	0.332	&	0.532	&	0.518	&	0.667	&	0.382	&	0.382	\\																
\hline
\addlinespace[0.1em]
\multicolumn{1}{c}{\textbf{Deep Learning} } & \multicolumn{1}{l}{} & \multicolumn{1}{l}{} & \multicolumn{1}{l}{} & \multicolumn{1}{l}{} & \multicolumn{1}{l}{} & \multicolumn{1}{l}{} & \multicolumn{1}{l}{} & \multicolumn{1}{l}{} & \multicolumn{1}{l}{} & \multicolumn{1}{l}{} & \multicolumn{1}{l}{} & \multicolumn{1}{l}{} \\ \hline	
\addlinespace[0.1em]

GDL	&	n/a	&	n/a	&	n/a	&	0.825	&	0.965	&	0.958	&	0.497	&	0.664	&	0.655	&	0.444	&	0.020	&	0.020	\\																					
RCC	&	0.876	&	0.893	&	0.893	&	0.831	&	0.963	&	0.957	&	0.484	&	0.85	&	0.836	&	0.356	&	0.138	&	0.138	\\																					
RCC-DR	&	0.698	&	0.827	&	0.828	&	0.825	&	0.963	&	0.957	&	0.579	&	0.882	&	0.874	&	0.676	&	0.442	&	0.442	\\																					
RCC-DR(SGD)	&	0.696	&	0.827	&	0.827	&	0.855	&	0.967	&	0.961	&	0.473	&	0.845	&	0.830	&	0.354	&	0.106	&	0.106	\\																					

DCN	&	0.560	&	0.570	&	0.570	&	0.6002	&	0.830	&	0.810	&	0.620	&	0.810	&	0.790	&	0.730	&	0.470	&	0.470	\\																					
DEC	&	0.867	&	0.853	&	0.840	&	0.815	&	0.645	&	0.611	&	0.643	&	0.811	&	0.807	&	0.683	&	0.504	&	0.500	\\																					
JULE	&	0.800	&	0.900	&	0.900	&	0.911	& 0.983	&	0.979	&	0.342	&	0.587	&	0.574	&	-	&	-	&	-	\\																					
DEPICT	& 0.968	&	0.919	&	0.919	&	0.402	&	0.678	&	0.667	&	0.586	&	0.790	&	0.785	&	-	&	-	&	-	\\																					
DCC	&	0.963	&	0.915	&	0.913	&	0.858	&	0.967	&	0.962	&	0.699	&	0.908	& 0.903	&	0.563	& 0.498 &	0.495	\\																					
\hline \hline	

\addlinespace[0.1em]


DMS ($subtract$)
& \textbf{	0.976}	&	\textbf{0.939}	&	\textbf{0.939}	&	\textbf{0.990}	&0.996	&	\textbf{0.994}	& \textbf{0.923}	&\textbf{	0.949}	&	\textbf{0.912}	&	\textbf{0.760}	&	0.587	&	\textbf{0.532}	\\		

DMS ($concat$)
& 0.958		& 0.914		& 0.896		&	\textbf{0.990}	&	\textbf{0.997}	&\textbf{	0.994}	& 0.832		&	0.896	& 0.820		& 0.582	& \textbf{0.603}	&	0.453	\\		


\hline																																				
\end{tabular}	


    \label{table:results}
    \end{table*}
    
\subsection{Ablation} 
\label{ablation} 

In this section we evaluate our training method and network architecture. We test the number of mean shift iterations used during training, the number of layers used in the kernel and the robustness of subtraction and concatenation based kernels to limited training data.

\noindent\textbf{Training Iterations.}
We test the number of iterations used during the training of our kernel, the results are shown in Table \ref{table:ablation}.
We hypothesise that when a small number of iterations are used the network does not have the sufficient flexibility to learn the nuance of the given task, however, with a large number of iterations the network is not incentivised to effectively define the clusters as it has many opportunities to improve its prediction. We found that the number of iterations required for convergence at inference time increased proportionally with both the number of training iterations and the complexity of the dataset. This verifies the idea that a single pass through the kernel is less able to distinguish inliers from outliers given more training iterations.
Given four training iterations the time taken to converge at inference time varied between five iterations and a single iteration depending on the complexity of the dataset.

\noindent\textbf{Kernel Depth.} We also tested the effect of changing the number of fully connected layers used in the differentiable kernel, as shown in Table \ref{table:ablation}.
All networks were trained for the same number of epochs, we used a sufficient number of epochs to ensure all networks had converged. 
After three layers the performance improvements are negligible, while training time and likelihood of overfitting increase. For this reason we use three fully connected layers for all other tests.

\noindent\textbf{Concatenation and Subtraction Kernels.}
We tested two differentiable kernels, the subtraction based kernel,
which takes the difference of the data point and sample mean, $(\mathbf{x} - \bar{\mathbf{x}}_i)$, and an alternate, concatenation based kernel,
which takes a 
concatenation of the data point and sample mean, $concat(\mathbf{x}, \bar{\mathbf{x}}_i)$.
The concatenation requires the addition of a convolutional layer, with a rectangular $ 2 \times 1$ kernel, before the fully connected layers. We use the same architecture as the original differential kernel after this convolution. 
Both architectures are outlined in Table \ref{table:Architecture}.

We hypothesised the use of a concatenation and a convolutional layer would be more effective than the simpler subtraction as a greater amount of information is used. It should be noted that the concatenation and a single convolutional layer could simply learn the subtraction function. The performances are of both are shown in Table \ref{table:results}.

\begin{figure}
 \centering
            \includegraphics[width=0.5\linewidth]{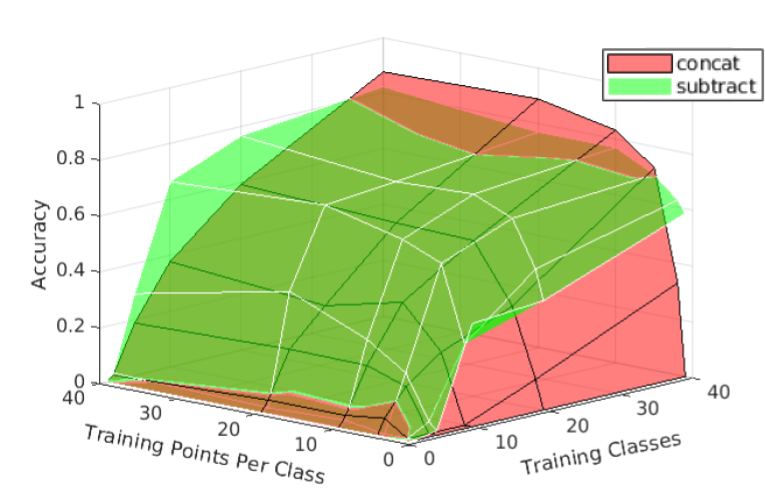}
            \caption{The effect of varying the distribution of training data on the performance of our subtraction and convolution based kernels.
             For this test we explicitly limit the number of classes and the maximum number of points per class with side information during training.
             This shows the methods ability to discover novel classes as there is very little decrease in performance of the subtraction based method when half of the classes are absent during training.}
    \label{fig:class_and_points}
 \end{figure}

We further test the performance of our subtraction and concatenation based kernels when less side information is available. 
Here we explicitly limit the number of classes and points per class with side information, this is unlike our other experiments where the points with side information are randomly selected.
The results of this experiment appear in Figure \ref{fig:class_and_points}. 

It can be seen that both kernels degrade similarly when fewer data points per class are seen during training with little performance effect unless the number of data points is significantly restricted. It can also be seen that that while the concatenation based kernel performs better when all ground truth classes are present in the training data, it's performance is significantly reduced when this is not the case.
The subtraction based kernel on the other hand, has little degradation in performance when decreasing the classes present during training, demonstrating its better ability to generalise to unseen data.
This could be due to the fact the subtraction based kernel, unlike the concatenation kernel is only presented with the relative position of points to the current sample mean. During training the subtraction based kernel is unable to use the global position of the training clusters as a crutch and must learn to find a better description of similarity.  This experiment also clearly shows the methods ability to correctly identify novel classes.

\section{Conclusion}\label{conclusion}
In this paper we have presented a new approach to clustering, one which learns to cluster points directly from side information in the form of pairwise similarity. Unlike previous methods, we do not require any other knowledge of the dataset as a whole; we do not need to know the number of clusters, their centers nor any kind of distance metric as a prior.
We demonstrate this approach using DMC, a deep clustering method inspired by the mean shift algorithm. 

DMS learns the requirements of a task and uses this to out perform state of the art deep learning based clustering approaches on both the intrinsic and non intrinsic tasks without the need to specify any task or dataset specific hyper-parameters.
This flexibility along with DMS's performance in high dimensional feature spaces and its lightweight nature enables it to be seamlessly incorporated into other deep learning applications for end-to-end training or as a head.
    \begin{table}
        \centering
        \caption{The performance of our the differentiable kernel with varied training iterations and number of fully connected layers. Both experiments were conducted on the YTF dataset.}
\setlength{\tabcolsep}{3pt}
\begin{tabular}{|c|cccc|cccc|}
\hline
 & \multicolumn{4}{c|}{\textbf{Training Iterations}} & \multicolumn{4}{c|}{\textbf{Fully Connected Layers}} \\
 & 2 & 4 & 6 & 8 & 1 & 2 & 3 & 4\\\hline
 
\textbf{ACC} & 0.818 & \textbf{0.923} & 0.890 & 0.780 & 0.069 & 0.920 & \textbf{0.923} & 0.905\\
\textbf{NMI} & 0.920 & \textbf{0.949} & 0.915 & 0.837 & 0.000 & 0.934 & 0.939 & \textbf{0.942}\\
\textbf{AMI} & 0.847 & \textbf{0.912} & 0.883 & 0.813 & 0.000 & 0.895 & \textbf{0.912} & 0.898\\

\hline
\end{tabular}






    \label{table:ablation}
    \end{table}

\newpage
\bibliographystyle{splncs}\bibliography{bibl}

\begin{thebibliography}{10}

\bibitem{Lloyd1982}
Lloyd, S.P.:
\newblock {Least Squares Quantization in PCM}.
\newblock IEEE Transactions on Information Theory \textbf{28} (1982)  129--137

\bibitem{Sander1998}
Sander, J., Ester, M., Kriegel, H.P., Xu, X.:
\newblock {Density-based clustering in spatial databases: The algorithm GDBSCAN
  and its applications}.
\newblock Data Mining and Knowledge Discovery \textbf{2} (1998)  169--194

\bibitem{Fukunaga1975}
Fukunaga, K., Hostetler, L.D.:
\newblock {The Estimation of the Gradient of a Density Function, with
  Applications in Pattern Recognition}.
\newblock IEEE Transactions on Information Theory \textbf{21} (1975)  32--40

\bibitem{Shah2018}
Shah, S.A., Koltun, V.:
\newblock {Robust continuous clustering}.
\newblock Proceedings of the National Academy of Sciences of the United States
  of America \textbf{114} (2017)  9814--9819

\bibitem{shah2018deep}
Shah, S.A., Koltun, V.:
\newblock Deep continuous clustering.
\newblock arXiv preprint arXiv:1803.01449 (2018)

\bibitem{Chang2017}
Chang, J., Wang, L., Meng, G., Xiang, S., Pan, C.:
\newblock {Deep Adaptive Image Clustering}.
\newblock In: Proceedings of the IEEE International Conference on Computer
  Vision. (2017)  5880--5888

\bibitem{Yang2017}
Yang, B., Fu, X., Sidiropoulos, N.D., Hong, M.:
\newblock {Towards K-means-friendly spaces: Simultaneous deep learning and
  clustering}.
\newblock In: 34th International Conference on Machine Learning. Volume~8.
  (2017)  5888--5901

\bibitem{Xing}
Xing, E.P., Ng, A.Y., Jordan, M.I., Russell, S.:
\newblock {Distance metric learning, with application to clustering with
  side-information}.
\newblock Advances in neural information processing systems. (2003)

\bibitem{han2020automatically}
Han, K., Rebuffi, S.A., Ehrhardt, S., Vedaldi, A., Zisserman, A.:
\newblock Automatically discovering and learning new visual categories with
  ranking statistics.
\newblock arXiv preprint arXiv:2002.05714 (2020)

\bibitem{Han}
Han, K., Vedaldi, A., Zisserman, A.:
\newblock Learning to discover novel visual categories via deep transfer
  clustering.
\newblock In: Proceedings of the IEEE International Conference on Computer
  Vision. (2019)  8401--8409

\bibitem{Dizaji2017}
Dizaji, K.G., Herandi, A., Deng, C., Cai, W., Huang, H.:
\newblock {Deep Clustering via Joint Convolutional Autoencoder Embedding and
  Relative Entropy Minimization}.
\newblock In: Proceedings of the IEEE International Conference on Computer
  Vision. (2017)  5747--5756

\bibitem{Hsu2015}
Hsu, Y.C., Kira, Z.:
\newblock {Neural network-based clustering using pairwise constraints}.
\newblock arXiv preprint, arXiv:1511.06321 (2015)

\bibitem{Liu2015}
Liu, H., Fu, Y.:
\newblock {Clustering with Partition Level Side Information}.
\newblock 2015 IEEE International Conference on Data Mining (2015)

\bibitem{Ceccarelli2008}
Ceccarelli, M., Maratea, A.:
\newblock {Improving fuzzy clustering of biological data by metric learning
  with side information}.
\newblock International Journal of Approximate Reasoning (2008)

\bibitem{Li2018}
Li, F., Qiao, H., Zhang, B.:
\newblock {Discriminatively boosted image clustering with fully convolutional
  auto-encoders}.
\newblock Pattern Recognition \textbf{83} (2018)  161--173

\bibitem{VonLuxburg2007}
{Von Luxburg}, U.:
\newblock {A tutorial on spectral clustering}.
\newblock Statistics and Computing \textbf{17} (2007)  395--416

\bibitem{Arthur2007}
Arthur, D., Vassilvitskii, S.:
\newblock {K-means++: The advantages of careful seeding}.
\newblock In: Proceedings of the Annual ACM-SIAM Symposium on Discrete
  Algorithms. (2007)  1027--1035

\bibitem{ailon2018approximate}
Ailon, N., Bhattacharya, A., Jaiswal, R.:
\newblock Approximate correlation clustering using same-cluster queries.
\newblock In: Latin American Symposium on Theoretical Informatics, Springer
  (2018)  14--27

\bibitem{choudhury2019top}
Choudhury, T., Shah, D., Karamchandani, N.:
\newblock Top-m clustering with a noisy oracle.
\newblock In: 2019 National Conference on Communications (NCC). (2019)  1--6

\bibitem{kim2017relaxed}
Kim, T., Ghosh, J.:
\newblock Relaxed oracles for semi-supervised clustering.
\newblock arXiv preprint arXiv:1711.07433 (2017)

\bibitem{mazumdar2017clustering}
Mazumdar, A., Saha, B.:
\newblock Clustering with noisy queries.
\newblock In: Advances in Neural Information Processing Systems. (2017)
  5788--5799

\bibitem{Murtagh2014}
Murtagh, F., Legendre, P.:
\newblock {Ward's Hierarchical Agglomerative Clustering Method: Which
  Algorithms Implement Ward's Criterion?}
\newblock Journal of Classification \textbf{31} (2014)  274--295

\bibitem{Zhang2012}
Zhang, W., Wang, X., Zhao, D., Tang, X.:
\newblock {Graph degree linkage: Agglomerative clustering on a directed graph}.
\newblock In: Lecture Notes in Computer Science (including subseries Lecture
  Notes in Artificial Intelligence and Lecture Notes in Bioinformatics). (2012)
   428--441

\bibitem{Nie2011}
Nie, F., Zeng, Z., Tsang, I.W., Xu, D., Zhang, C.:
\newblock {Spectral embedded clustering: A framework for in-sample and
  out-of-sample spectral clustering}.
\newblock IEEE Transactions on Neural Networks \textbf{22} (2011)  1796--1808

\bibitem{Shi}
Shi, J., Malik, J.:
\newblock {Normalized cuts and image segmentation}.
\newblock Technical Report~8, IEEE Trans. Pattern Anal. Machine Intell.(PAMI).
  (2000)

\bibitem{Aggarwal2001}
Aggarwal, C.C., Hinneburg, A., Keim, D.A.:
\newblock {On the surprising behavior of distance metrics in high dimensional
  space}.
\newblock In: Lecture Notes in Computer Science (including subseries Lecture
  Notes in Artificial Intelligence and Lecture Notes in Bioinformatics). Volume
  1973.
\newblock ICDT (2001)  420--434

\bibitem{Domingos:2012:FUT:2347736.2347755}
Domingos, P.:
\newblock {A few useful things to know about machine learning}.
\newblock Communications of the ACM \textbf{55} (2012)  78--87

\bibitem{Bellman1962}
Wright, E.M., Bellman, R.:
\newblock {Adaptive Control Processes: A Guided Tour}.
\newblock The Mathematical Gazette \textbf{46} (1962)  160

\bibitem{DBLP:journals/corr/abs-1801-07648}
Aljalbout, E., Golkov, V., Siddiqui, Y., Strobel, M., Cremers, D.:
\newblock {Clustering with Deep Learning: Taxonomy and New Methods}.
\newblock CoRR \textbf{abs/1801.0} (2018)

\bibitem{Yang2016}
Yang, J., Parikh, D., Batra, D.:
\newblock {Joint unsupervised learning of deep representations and image
  clusters}.
\newblock In: Proceedings of the IEEE Computer Society Conference on Computer
  Vision and Pattern Recognition. (2016)  5147--5156

\bibitem{Xie2016}
Xie, J., Girshick, R., Farhadi, A.:
\newblock {Unsupervised deep embedding for clustering analysis}.
\newblock In: 33rd International Conference on Machine Learning. (2016)
  740--749

\bibitem{Hu2017}
Hu, W., Miyato, T., Tokui, S., Matsumoto, E., Sugiyama, M.:
\newblock {Learning discrete representations via information maximizing
  self-augmented training}.
\newblock In: 34th International Conference on Machine Learning. (2017)
  2467--2481

\bibitem{Hsu2018}
Hsu, C.C., Lin, C.W.:
\newblock {CNN-Based joint clustering and representation learning with feature
  drift compensation for large-scale image data}.
\newblock IEEE Transactions on Multimedia \textbf{20} (2018)  421--429

\bibitem{VittalPremachandran2017}
{Vittal Premachandran}, Yuille, A.L.:
\newblock {Unsupervised Learning Using Generative Adversarial Training and
  Clustering}.
\newblock The International Conference on Learning Representations (2017)

\bibitem{Fogel2019}
Fogel, S., Averbuch-Elor, H., Cohen-Or, D., Goldberger, J.:
\newblock {Clustering-Driven Deep Embedding With Pairwise Constraints}.
\newblock IEEE Computer Graphics and Applications (2019)

\bibitem{Kong2018}
Kong, S., Fowlkes, C.:
\newblock {Recurrent Pixel Embedding for Instance Grouping}.
\newblock In: Proceedings of the IEEE Computer Society Conference on Computer
  Vision and Pattern Recognition. (2018)  9018--9028

\bibitem{unknown}
Shukla, A., Cheema, G.S., Anand, S.:
\newblock {Semi-Supervised Clustering with Neural Networks} (2018)

\bibitem{Mazumdar2017}
Mazumdar, A., Saha, B.:
\newblock {Query complexity of clustering with side information}.
\newblock In: Advances in Neural Information Processing Systems. (2017)

\bibitem{Cheng1995}
Cheng, Y.:
\newblock {Mean Shift, Mode Seeking, and Clustering}.
\newblock IEEE Transactions on Pattern Analysis and Machine Intelligence
  \textbf{17} (1995)  790--799

\bibitem{LeCun1998}
LeCun, Y., Bottou, L., Bengio, Y., Haffner, P.:
\newblock {Gradient-based learning applied to document recognition}.
\newblock Proceedings of the IEEE \textbf{86} (1998)  2278--2323

\bibitem{Nene1996}
Nene, S., Nayar, S., Murase, H.:
\newblock {Columbia Object Image Library (COIL-20)}.
\newblock Technical Report \textbf{95} (1996)  223--303

\bibitem{Wolf2011}
Wolf, L., Hassner, T., Maoz, I.:
\newblock {Face recognition in unconstrained videos with matched background
  similarity}.
\newblock In: Proceedings of the IEEE Computer Society Conference on Computer
  Vision and Pattern Recognition. (2011)  529--534

\bibitem{Lewis2004}
Lewis, D.D., Yang, Y., Rose, T.G., Li, F.:
\newblock {RCV1: A new benchmark collection for text categorization research}.
\newblock Journal of Machine Learning Research \textbf{5} (2004)  361--397

\bibitem{Yang2010}
Yang, Y., Xu, D., Nie, F., Yan, S., Zhuang, Y.:
\newblock {Image clustering using local discriminant models and global
  integration}.
\newblock IEEE Transactions on Image Processing \textbf{19} (2010)  2761--2773

\bibitem{Strehl2002}
Strehl, A., Ghosh, J.:
\newblock {Cluster ensembles - A knowledge reuse framework for combining
  multiple partitions}.
\newblock Journal of Machine Learning Research \textbf{3} (2003)  583--617

\bibitem{Vinh:2010:ITM:1756006.1953024}
Vinh, N.X., Epps, J., Bailey, J.:
\newblock {Information theoretic measures for clusterings comparison: Variants,
  properties, normalization and correction for chance}.
\newblock Journal of Machine Learning Research \textbf{11} (2010)  2837--2854

\bibitem{Kuhn1955}
Kuhn, H.W.:
\newblock {The Hungarian method for the assignment problem}.
\newblock 50 Years of Integer Programming 1958-2008: From the Early Years to
  the State-of-the-Art \textbf{2} (2010)  29--47

\end{thebibliography}
\end{document}